\documentclass[11pt,a4paper]{article}
\usepackage{acl2015}
\usepackage{times}
\usepackage{url}
\usepackage{latexsym}
\usepackage{amsmath,bm}
\usepackage{amsfonts}
\usepackage{mathrsfs}
\usepackage{amsbsy}
\usepackage{graphicx}
\usepackage{xcolor}
\usepackage{multirow}
\usepackage{paralist}
\usepackage{subfigure}
\newcommand{\tabincell}[2]{\begin{tabular}{@{}#1@{}}#2\end{tabular}}
\usepackage{latexsym}

\title{Discriminative Neural Sentence Modeling\\
 by Tree-Based Convolution}

\author{Lili Mou\thanks{$\ \ \ \ $These authors contribute equally to this paper.},
Hao Peng$^{*}\!\!$,
Ge Li\thanks{$\ \ \ \ $Corresponding author.}, Yan Xu, Lu Zhang, Zhi Jin\\
\{moull12, lige, xuyan14, zhanglu, zhijin\}@sei.pku.edu.cn\\
penghao.pku@gmail.com\\
Software Institute, Peking University, 100871, P. R. China
}

\date{}

\begin{document}
\maketitle


\begin{abstract}
This paper proposes a tree-based convolutional neural network (TBCNN)
for discriminative sentence modeling.
Our models leverage either constituency trees or
dependency trees of sentences.
The tree-based convolution process extracts sentences' structural features, and
these features are aggregated by max pooling.
Such architecture allows short propagation paths between the output layer and 
underlying feature detectors, which enables
effective structural feature learning and extraction.
We evaluate our models on two tasks: sentiment analysis and question classification.
In both experiments, TBCNN outperforms previous state-of-the-art results,
including existing neural networks and dedicated feature/rule engineering.
We also make efforts to visualize the tree-based convolution process,
shedding light on how our models work.
\end{abstract}

\section{Introduction}
\begin{figure*}[!t]
\centering
\vspace{-.2cm}

\includegraphics[width=.93\textwidth]{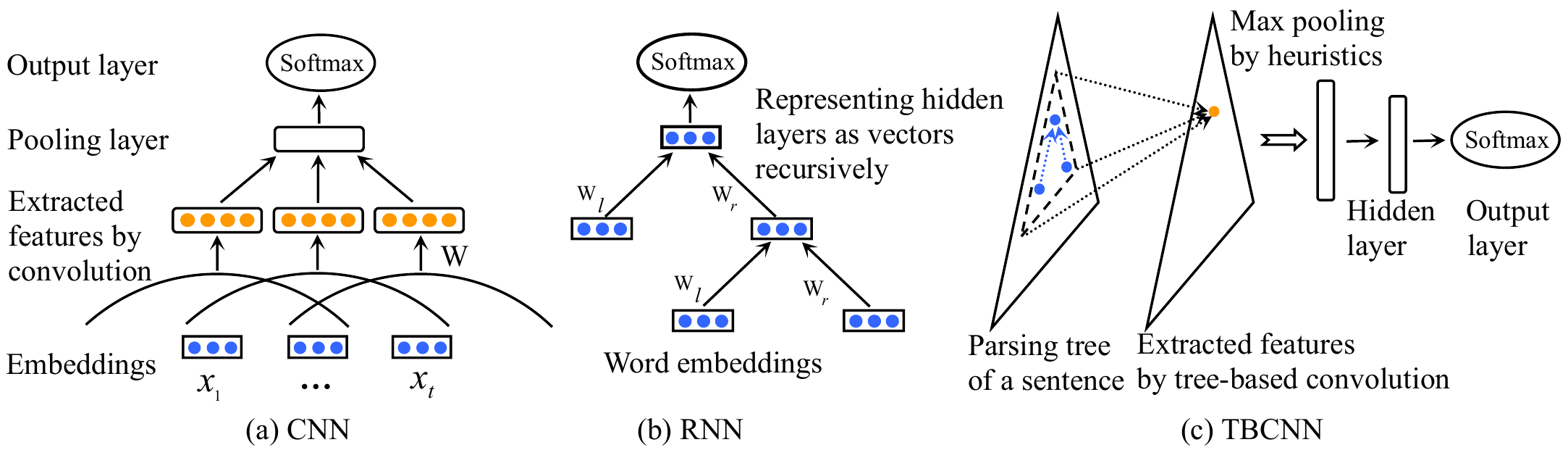}

\vspace{-.5cm}
\caption{A comparison of information flow in the convolutional neural network (CNN), the
recursive neural network (RNN), and the tree-based convolutional neural network (TBCNN).
}\label{fCompare}

\vspace{-.3cm}
\end{figure*}

Discriminative sentence modeling aims to capture sentence meanings,
and classify sentences according to certain criteria (e.g., sentiment).
It is related to various tasks of interest,
and has attracted much attention in the NLP community \cite{classify_sen,classify_sen_2,AdaCNN}.
Feature engineering---for example, $n$-gram features \cite{ngram}, 
dependency subtree features \cite{dependencyfeature}, or more dedicated ones \cite{QC}---can
play an important role in modeling sentences.
Kernel machines, e.g., SVM, are exploited in 
\newcite{kernel} and \newcite{kernel2} by
specifying a certain measure of similarity between sentences, without
explicit feature representation.


Recent advances of neural networks bring new techniques in understanding natural languages, 
and have exhibited considerable potential.
\newcite{LM} and \newcite{word2vec} propose
unsupervised approaches to learn word embeddings,
mapping discrete words to real-valued vectors in a meaning space.
\newcite{paravec} extend such approaches to learn sentences' and paragraphs'
representations.
Compared with human engineering, 
neural networks serve as a way of automatic feature learning \cite{RL}.

Two widely used neural sentence models are
convolutional neural networks (CNNs) 
and recursive neural networks (RNNs). 
CNNs can extract words' 
neighboring features effectively with short propagation paths,
but they do not capture inherent sentence structures (e.g., parsing trees).
RNNs encode, to some extent, structural information by 
recursive semantic composition 
along a parsing tree. 
However, they may have difficulties in learning deep dependencies because of
long propagation paths \cite{difficulty}.
(CNNs/RNNs and a variant, recurrent networks, will be reviewed in Section \ref{sBackground}.)

A curious question is whether we can combine the advantages of CNNs and RNNs,
i.e., whether we can exploit sentence structures (like RNNs)
effectively with short propagation paths (like CNNs).

In this paper, we propose a novel neural architecture for discriminative sentence modeling,
called the \textit{Tree-Based Convolutional Neural Network} (TBCNN).
Our models can leverage different sentence parsing trees, e.g.,
constituency trees and dependency trees. The model variants are denoted as c-TBCNN and d-TBCNN, respectively.
The idea of tree-based convolution is to apply a set of subtree feature detectors,
sliding over the entire parsing tree of a sentence; 
then pooling aggregates these extracted feature vectors by taking the maximum value in each dimension.
One merit of such architecture is that all features, along the tree, have short propagation paths
to the output layer, and hence structural information can be learned effectively.

TBCNNs are evaluated on two tasks, sentiment analysis and question classification;
our models have outperformed previous state-of-the-art results in both experiments.
To understand how TBCNNs work, we also visualize the network by plotting the convolution process.
We make our code and results available on our project website.\footnote{
https://sites.google.com/site/tbcnnsentence/ (This site is properly anonymized, and complies with the double-blind review requirement.)}

\section{Background and Related Work}\label{sBackground}

In this section, we present the background and related work regarding
two prevailing neural architectures for discriminative sentence modeling.

\subsection{Convolutional Neural Networks}

Convolutional neural networks (CNNs), early used for image processing \cite{lenet},
turn out to be effective with natural languages as well.
Figure \ref{fCompare}a depicts a classic convolution
process on a sentence \cite{unified}.
A set of fixed-width-window feature detectors 
slide over the sentence, and output the extracted features.
Let $t$ be the window size, and $\bm x_1, \cdots, \bm x_t \in\mathbb{R}^{n_e}$
be $n_e$-dimensional
word embeddings.
The output of convolution, evaluated at the current position, is

\vspace{-.4cm}
$$\bm y=f\left(W\cdot \left[
\bm x_1;\cdots;\bm x_t\right]+\bm b\right)$$

\vspace{-.1cm}
\noindent where $\bm y\in \mathbb{R}^{n_c}$ ($n_c$ is the number of feature detectors). $W\in\mathbb{R}^{n_c\times (t\cdot n_e)}$ and $\bm b\in\mathbb{R}^{n_c}$ are parameters; $f$ is the activation function. Semicolons represent
 column vector concatenation.
After convolution, the extracted features are pooled to a fixed-size vector for classification.

Convolution can extract neighboring information effectively.
However, the features are ``local''---words that are not in a same convolution window do not interact with each other, even though they may be semantically related.
\newcite{CNNNLP} build deep convolutional networks so that
local features can mix at high-level layers.
Similar deep CNNs include \newcite{cnn2} and \newcite{cnn3}.
All these models are ``flat,'' by which we mean no structural information
is used explicitly.


\subsection{Recursive Neural Networks}

Recursive neural networks (RNNs), proposed in \newcite{RAE},
utilize sentence parsing trees.
In the original version, RNN is built upon a binarized constituency tree.
Leaf nodes correspond to words in a sentence, represented by
$n_e$-dimensional embeddings.
Non-leaf nodes are sentence constituents,
coded by child nodes recursively.
Let node $p$ be the parent of $c_1$ and $c_2$,
 vector representations denoted as
$\bm p$, $\bm c_1$, and $\bm c_2$. The parent's representation is composited by

\vspace{-.4cm}
\begin{equation}
\bm p=f(W\cdot[\bm c_1; \bm c_2]+\bm b)\label{eRAE}
\end{equation}

\vspace{-.1cm}

\noindent where $W$ and $\bm b$ are parameters.
This process is done recursively along the tree; the root vector is then used for supervised classification (Figure \ref{fCompare}b).

Dependency parsing and the combinatory categorical grammar can also be exploited
as RNNs' skeletons \cite{ccg,dependencyrnn}.
\newcite{deepRNN} build deep RNNs to enhance information interaction.
Improvements for semantic compositionality include matrix-vector interaction \cite{matrixvector}, tensor interaction \cite{RNN}. They are more suitable for 
capturing logical information in sentences, such as negation and exclamation.

One potential problem of RNNs is that the long propagation paths---through which leaf nodes are connected to the output layer---may lead to information loss.
Thus, RNNs bury illuminating information under a complicated neural architecture.
Further, during back-propagation over a long path, gradients tend to vanish (or blow up),
which makes training difficult \cite{difficulty}.
Long short term memory (LSTM), first proposed for modeling
time-series data \cite{lstm}, is integrated to RNNs to alleviate this problem \cite{lstm1,lstm2,lstm3}.

\textbf{Recurrent networks}.
A variant class of RNNs is the recurrent neural network \cite{rnndifficult,responding},
whose architecture is a rightmost tree.
In such models, meaningful tree structures are also lost, similar to CNNs.

\section{Tree-based Convolution}

This section introduces the proposed
tree-based convolutional neural networks (TBCNNs).
Figure \ref{fCompare}c depicts the convolution process on a tree.

First, a sentence is converted to a parsing tree,
either a constituency or dependency tree.
The corresponding model variants are denoted as c-TBCNN and d-TBCNN.
Each node in the tree is represented as a distributed, real-valued vector.

Then, we design a set of fixed-depth subtree feature detectors, called the
\textit{tree-based convolution window}.
The window
slides over the entire tree to extract structural information
of the sentence, illustrated by a dashed triangle in Figure \ref{fCompare}c.
Formally, let us assume we 
have $t$ nodes in the convolution window, $\bm x_1,\cdots, \bm x_t$, 
each represented as an $n_e$-dimensional vector.
Let $n_c$ be the number of feature detectors.
The output of the tree-based convolution window, evaluated at the current subtree, is
given by the following generic equation.

\vspace{-.3cm}
\begin{equation}\label{eConv}
\bm y=f\left(\sum_{i=1}^tW_i\!\cdot\!\bm x_i+\bm b\right)
\end{equation}
\vspace{-.3cm}

\begin{figure*}[!t]
\vspace{-.2cm}

\centering
\includegraphics[width=.9\textwidth]{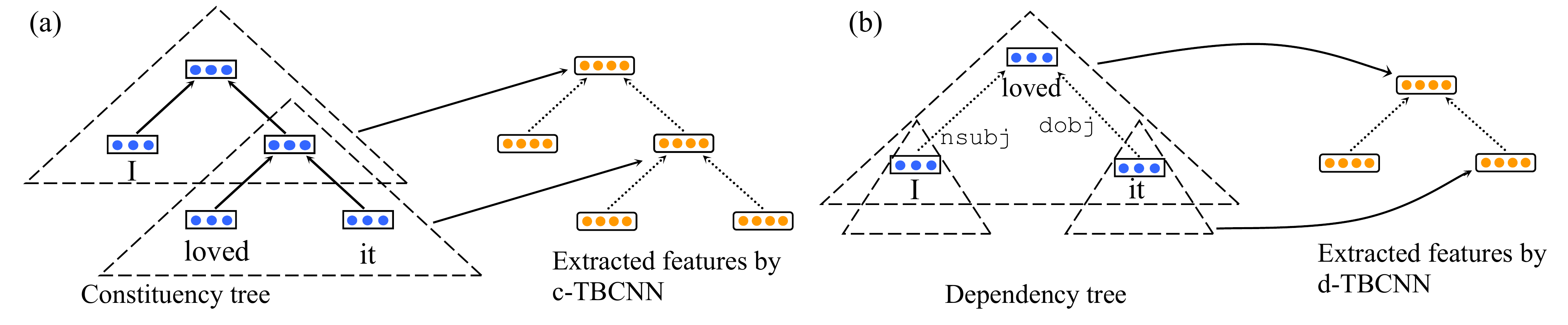}

\vspace{-.5cm}
\caption{Tree-based convolution in (a) c-TBCNN, and
(b) d-TBCNN.
The parsing trees correspond to the sentence ``I loved it.''
The dashed triangles illustrate a shared-weight convolution
window sliding over the tree. For clarity, only two positions are drawn in c-TBCNN.
Notice that
dotted arrows are not part of neural connections; they merely indicate
the topologies of tree structures. Specially, an edge 
$a\overset{r}{\rightarrow} b$
in the dependency tree refers to $a$ being governed by $b$ with dependency type $r$.}
\vspace{-.35cm}
\label{fTBCNN}
\end{figure*}

\noindent where $W_i\in\mathbb{R}^{n_e\times n_c}$ is the weight parameter associated with
node $x_i$; $\bm b\in\mathbb{R}^{n_c}$ is the bias term.

Extracted features are thereafter packed into one or more fixed-size vectors by max pooling, that is, the maximum value in each dimension is taken.
Finally, we add a fully connected hidden layer, and a $\operatorname{softmax}$ output layer.

From the designed architecture (Figure \ref{fCompare}c), we see that our TBCNN models allow short
propagation paths between the output layer and any position in the tree.
Therefore structural feature learning becomes effective.

Several main technical points in tree-based convolution include:
(1) How can we represent hidden nodes as vectors in constituency trees?
(2) How can we determine weights, $W_i$, for dependency trees, where nodes
may have different numbers of children?
(3) How can we pool varying sized and shaped features to fixed-size vectors?

In the rest of this section, we explain model variants in detail.
Particularly, Subsections \ref{sscTBCNN} and \ref{ssdTBCNN} address the first and second problems;
Subsection \ref{ssPool} deals with the third problem by introducing several pooling heuristics.
Subsection \ref{ssObjective} presents our training objective.
\subsection{c-TBCNN}\label{sscTBCNN}

Figure \ref{fTBCNN}a illustrates an example of the constituency tree, where
leaf nodes are words in the sentence, and
non-leaf nodes represent a grammatical constituent,
e.g., a noun phrase. Sentences are parsed by the Stanford parser;\footnote{
http://nlp.stanford.edu/software/lex-parser.shtml} further,
constituency trees are binarized for simplicity.

One problem of constituency trees is that non-leaf nodes do not have such vector representations as word embeddings.
Our strategy is to pretrain the constituency tree with an RNN by Equation \ref{eRAE} \cite{RAE}.
After pretraining, vector representations of nodes are fixed.

We now consider the tree-based convolution process in c-TBCNN with a two-layer-subtree
convolution window,
which operates on a parent node $p$ and its direct children $c_l$ and $c_r$, their vector representations denoted as $\bm p$, $\bm c_l$, and $\bm c_r$.
The convolution equation, specific for c-TBCNN, is

\vspace{-.3cm}
$$\bm y=f\left(W_p^{(c)}\!\cdot\!\bm p + W_l^{(c)}\!\cdot\!\bm c_l+W_r^{(c)}\!\cdot\!\bm c_r+\bm b^{(c)}\right)$$

\vspace{-.1cm}

\noindent where
$W_p^{(c)}$, $W_l^{(c)}$, and $W_r^{(c)}$ are weights associated with
the parent and its child nodes.
Superscript $(c)$ indicates that the weights are for c-TBCNN.
For leaf nodes, which do not have children,
we set $\bm c_l$ and $\bm c_r$ to be $\bm 0$.

Tree-based convolution windows can be extended to arbitrary depths straightforwardly.
The complexity is exponential to the depth of the window,
but linear to the number of nodes.
Hence, tree-based convolution, compared with ``flat'' CNNs,
does not add to computational cost, provided the same amount of information to process
at a time. In our experiments, we use convolution windows of depth 2.

\subsection{d-TBCNN}\label{ssdTBCNN}

Dependency trees are another representation of sentence structures.
The nature of dependency representation
leads to d-TBCNN's major difference from traditional convolution:
there exist nodes with different numbers of child nodes.
This causes trouble if we associate weight parameters according to
positions in the window, which is standard for traditional convolution, e.g.,
\newcite{unified} or c-TBCNN.

To overcome the problem, we extend the notion of convolution by
assigning weights according to dependency types
(e.g, \verb|nsubj|) rather than positions.
We believe this strategy makes much sense because dependency types \cite{dependencyparsing} 
reflect the relationship between a governing word and its child words.
To be concrete, the generic convolution formula (Equation \ref{eConv}) for d-TBCNN becomes

\vspace{-.5cm}
$$\bm y = f\left(W^{(d)}_p\!\cdot\!\textsl{}\bm p+\sum_{i=1}^n W^{(d)}_{r[c_i]}\!\cdot\!\bm c_i+\bm b^{(d)}\right)$$

\vspace{-.3cm}

\noindent where
$W^{(d)}_p$ is the weight parameter for the parent $ p$
(governing word);
$W^{(d)}_{r[c_i]}$ is the weight for child $c_i$,
who has grammatical relationship $r[c_i]$ to its parent, $p$. Superscript $(d)$ indicates the parameters are for d-TBCNN.
Note that we keep 15 most frequently occurred dependency types\textsl{};
others appearing rarely in the corpus are mapped to one shared weight matrix.

Both c-TBCNN and d-TBCNN have their own advantages:
d-TBCNN exploits structural features more efficiently
because of the compact expressiveness of dependency trees;
c-TBCNN may be more effective in integrating global features due to the underneath
pretrained RNN.

\subsection{Pooling Heuristics}\label{ssPool}
\begin{table*}
\small
\centering

\vspace{-.2cm}
\begin{tabular}{clr}
\hline
\hline
Task               & Data samples& Label\\
\hline
\multirow{3}{*}{\tabincell{c}{Sentiment\\ Analysis}}
     & Offers that rare combination of entertainment and education.
     & $++$
 \\
     & An idealistic love story that brings out the latent 15-year-old romantic in everyone. &$+$\\
     & Its mysteries are transparently obvious, and it's too slowly paced to be a thriller. &$-$\
    \\
\hline
\multirow{2}{*}{\tabincell{c}{Question\\ Classification}}
& What is the temperature at the center of the earth?  &
\verb"number"\\
      & What state did the Battle of Bighorn take place in? &
\verb"location"\\
\hline
\hline
\end{tabular}

\vspace{-.2cm}
\caption{Data samples in sentiment analysis and question classification.
In the first task, ``$++$'' refers to strongly positive; ``$+$'' and ``$-$'' refer to positive and negative, respectively.}\label{tExample}
\vspace{-.4cm}
\end{table*}

\begin{figure}[!t]
\centering
\vspace{-.25cm}

\includegraphics[width=.42\textwidth]{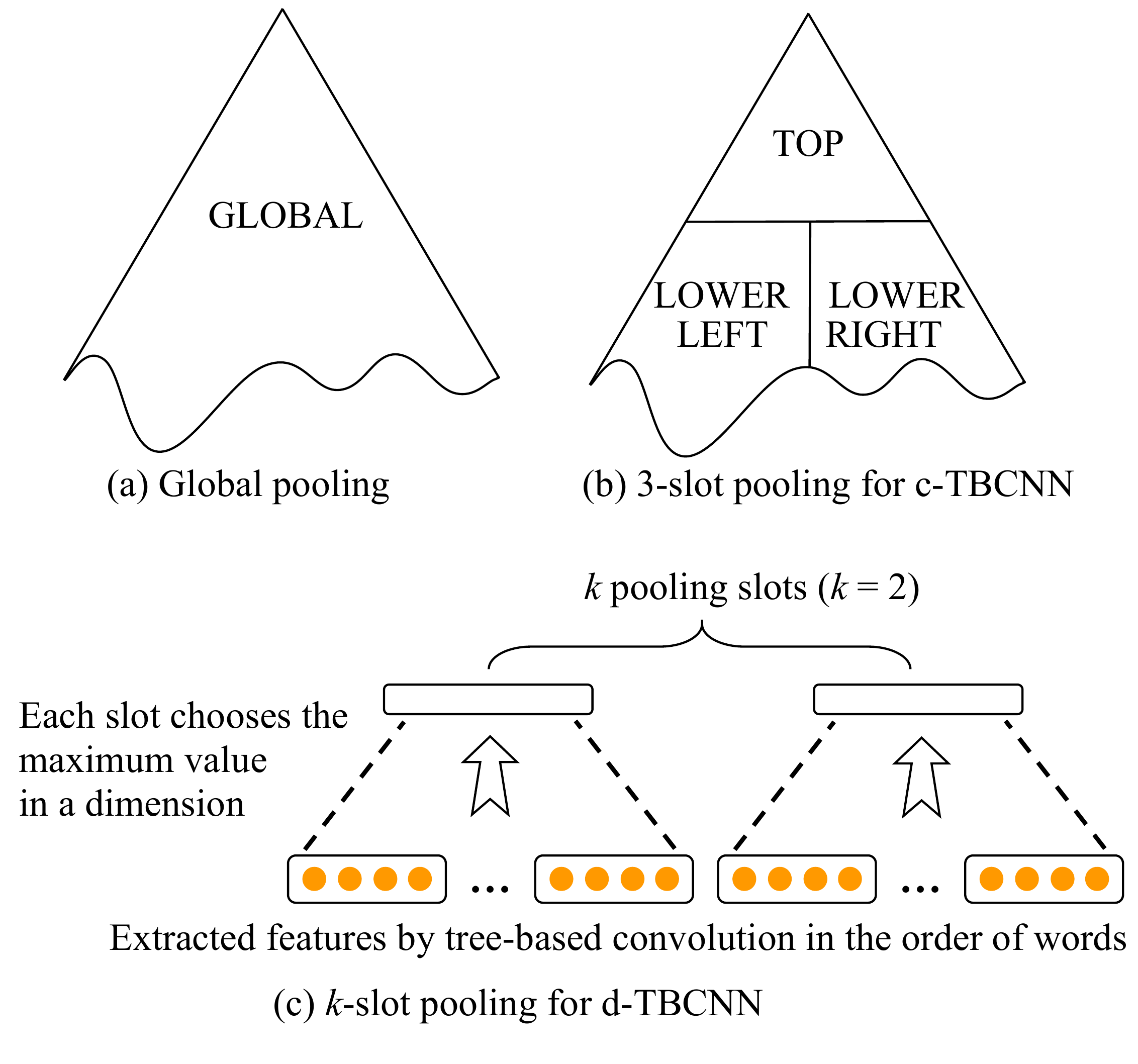}

\vspace{-.6cm}
\caption{Pooling heuristics. (a) Global pooling.
(b) 3-slot pooling for c-TBCNN. (c) $k$-slot pooling for d-TBCNN.}
\vspace{-.4cm}
\label{fPool}
\end{figure}

As different sentences may have different lengths and tree structures,
the extracted features by tree-based convolution also have topologies varying in size and shape.
Dynamic pooling \cite{dynamic} is a common technique for dealing with this problem.
We propose several heuristics for pooling along a tree structure.
Our generic design criteria for pooling include: (1) Nodes that are pooled to one slot should be ``neighboring'' from some viewpoint. (2) Each slot should have similar numbers of nodes, in expectation, that are pooled to it.
Thus, (approximately) equal amount of information is aggregated
along different parts of the tree. 
Following the above intuition, we propose pooling heuristics as follows.

\begin{compactitem}
\item Global pooling. All features are pooled to one vector, shown in Figure \ref{fPool}a.
We take
the maximum value in each dimension. This simple heuristic is applicable to
any structure, including c-TBCNN and d-TBCNN.

\item $3$-slot pooling for c-TBCNN. To preserve more information over different parts of
constituency trees, we propose 3-slot pooling (Figure \ref{fPool}b).
If a tree has maximum depth $d$,
we pool nodes of less than $\alpha\cdot d$ layers to a TOP slot ($\alpha$ is set to 0.6); lower nodes are pooled to slots LOWER\_LEFT
or LOWER\_RIGHT according to their relative position with respect to the root node.

For a constituency tree,
it is not completely obvious how to pool
features to more than 3 slots
and comply with the aforementioned criteria at the same time.
Therefore, we regard 3-slot pooling for c-TBCNN is a ``hard mechanism''
temporarily. Further improvement can be addressed in future work.

\item $k$-slot pooling for d-TBCNN. Different from constituency trees,
nodes in dependency trees are one-one corresponding to
words in a sentence.
Thus, a total order on features (after convolution) can be defined according to
their corresponding word orders.
For $k$-slot pooling,
we can adopt an ``equal allocation'' strategy, shown in Figure \ref{fPool}c.
Let $i$ be the position of a word in a sentence ($i= 1, 2, \cdots, n$).
Its extracted feature vector is pooled to the $j$-th slot, if

\vspace{-.3cm}

$$(j-1)\dfrac{\,n\,}k \le i \le j\dfrac{\, n\,}k$$
\end{compactitem}

\vspace{-.1cm}

We assess the efficacy of pooling quantitatively in Section \ref{sssPool}.
As we shall see by the experimental results, 
complicated pooling methods do preserve more information along tree structures to some
extent, but the effect is not large. TBCNNs are not very sensitive to pooling methods.
\subsection{Training Objective}\label{ssObjective}

After pooling, information is packed into one or more fixed-size vectors (slots).
We add a hidden layer, and then a $\operatorname{softmax}$
layer to predict the probability of each target label in a classification task.
The error function of a sample is the standard cross entropy loss, i.e.,
$J=-\sum_{i=1}^c t_i\log y_i$,
where $\bm t$ is the ground truth (one-hot represented),
$\bm y$ the output by $\operatorname{softmax}$, and $c$ the number of classes.
To regularize our model, we apply both $\ell_2$ penalty
and dropout \cite{dropout}.
Training details are further presented in Section \ref{ssSentiment} and \ref{ssQuestion}.

\section{Experimental Results}

In this section, we evaluate our models with two tasks,
sentiment analysis and question classification.
We also conduct quantitative and qualitative model analysis
in Subsection \ref{ssModel}.

\subsection{Sentiment Analysis}\label{ssSentiment}

\subsubsection{The Task and Dataset}
Sentiment analysis is a widely studied task for discriminative sentence modeling.
The Stanford sentiment treebank\footnote{http://nlp.stanford.edu/sentiment/} consists of more than 10,000 movie reviews.
Two settings are considered for sentiment prediction:
(1) fine-grained classification with 5 labels
(\verb|strongly positive|, \verb|positive|, \verb|neutral|,
\verb|negative|, and \verb|strongly| \verb|negative|), and
(2) coarse-gained polarity classification with 2 labels (\verb|positive| versus
\verb|negative|).
Some examples are shown in Table \ref{tExample}.
We use the standard split for training, validating, and testing, containing 8544/1101/2210 sentences for 5-class prediction.
Binary classification does not contain the \verb|neutral| class.

In the dataset, phrases (sub-sentences) are also tagged with sentiment labels.
RNNs deal with them naturally during the recursive process.
We regard sub-sentences as individual samples during training, like \newcite{CNNNLP} and \newcite{paravec}.
The training set therefore has more than 150,000 entries in total.
 For validating and testing, only whole sentences (root labels) are considered
 in our experiments.

Both c-TBCNN and d-TBCNN use the Stanford parser for data preprocessing.

\begin{table*}
\centering
\small
\begin{tabular}{c|c|c|c|l}
\hline
\hline
\textbf{Group} & \textbf{Method} &\textbf{5-class accuracy} & \textbf{2-class accuracy}&\textbf{Reported in}\\
\hline
\multirow{2}{*}{Baseline}  & SVM       &    40.7  & 79.4&    \newcite{RNN}   \\
                       & Na\"ive Bayes &    41.0  & 81.8&  \newcite{RNN}    \\
\hline
\multirow{4}{*}{CNNs} & 1-layer convolution  & 37.4  & 77.1  &   \newcite{CNNNLP}  \\
                      & Deep CNN             & 48.5  & 86.8  &  \newcite{CNNNLP}  \\
                      & Non-static		& 48.0   & 87.2      & \newcite{cnn2}\\
                      & Multichannel     & 47.4  & \textbf{88.1} & \newcite{cnn2}\\
\hline		
\multirow{6}{*}{RNNs} & Basic    &   43.2  & 82.4  &  \newcite{RNN}  \\
                      & Matrix-vector      &   44.4  & 82.9  &  \newcite{RNN}  \\
                      & Tensor             &   45.7  & 85.4  &  \newcite{RNN}  \\
                      & Tree LSTM (variant 1)			   &   48.0  & --    & \newcite{lstm3} \\
                      & Tree LSTM (variant 2)              &  50.6  & 86.9  &  \newcite{lstm1} \\
                      & Tree LSTM (variant 3) & 49.9 & 88.0 & \newcite{lstm2}\\
                      & Deep RNN            &  49.8  &\ \  86.6$^{\dag}$ &  \newcite{deepRNN}\\
\hline
\multirow{2}{*}{Recurrent} & LSTM           & 45.8       & 86.7  &\newcite{lstm1}\\
                           & bi-LSTM        & 49.1       & 86.8  &\newcite{lstm1}\\
\hline
\multirow{2}{*}{Vector} & Word vector avg.  &  32.7    & 80.1 & \newcite{RNN} \\
                        &  Paragraph vector &  48.7  & 87.8 & \newcite{paravec}\\
\hline
\multirow{2}{*}{TBCNNs}  &   c-TBCNN & 50.4          &\ \ 86.8$^\dag$  &  Our implementation\\
                        &   d-TBCNN  & \textbf{51.4} &\ \ 87.9$^\dag$  & Our implementation\\
\hline
\hline
\end{tabular}

\vspace{-.2cm}
\caption{Accuracy of sentiment prediction (in percentage).
For 2-class prediction, ``$\dag$'' remarks indicate that the network
is transferred directly from that of
5-class. 
}\label{tAcc}

\vspace{-.3cm}
\end{table*}

\subsubsection{Training Details}\label{ssHyperparameter}
This subsection describes training details for d-TBCNN,
where hyperparameters are chosen by validation.
c-TBCNN is mostly tuned synchronously (e.g., optimization algorithm, activation function)
with some changes in hyperparameters. c-TBCNN's settings can be found on our
(anonymized) website.

In our d-TBCNN model, the number of units is 300 for convolution and 200 for the last hidden layer.
Word embeddings are 300 dimensional,
pretrained ourselves using \verb|word2vec| \cite{word2vec} on the English Wikipedia corpus.
2-slot pooling is applied for d-TBCNN. (c-TBCNN uses 3-slot pooling.)

To train our model, we compute gradient by back-propagation
and apply stochastic gradient descent with mini-batch 200.
We use $\operatorname{ReLU}$ \cite{relu} as the activation function .

For regularization, we add $\ell_2$ penalty for weights
with a coefficient of $10^ {-5}$.
Dropout \cite{dropout}
is further applied to both weights and embeddings.
All hidden layers are dropped out by 50\%, and
embeddings 40\%.

\subsubsection{Performance}
Table \ref{tAcc} compares our models to state-of-the-art results in the task of sentiment analysis.
For 5-class prediction, d-TBCNN yields 51.4\% accuracy, outperforming the previous state-of-the-art result,
achieved by the RNN based on long-short term memory \cite{lstm1}.
c-TBCNN is slightly worse.
It achieves 50.4\% accuracy, ranking third in the
state-of-the-art list (including our d-TBCNN model).

Regarding 2-class prediction, we adopted a simple strategy in \newcite{deepRNN},\footnote{
Richard Socher, who first applies neural networks to this task,
thinks direct transfer is fine for binary classification. 
We followed this strategy for simplicity
as it is non-trivial to deal with the neutral sub-sentences in the training set 
if we train a  separate model. 
Our website reviews some related work and provides more discussions.}
where the 5-class network is ``transferred'' directly for binary classification,
with estimated target probabilities
(by 5-way $\operatorname{softmax}$) reinterpreted for 2 classes.
(The \verb|neutral| class is discarded as in other studies.)
This strategy enables us to take a glance at the stability of our TBCNN models, but places
itself in a difficult position. Nonetheless, our d-TBCNN model achieves 87.9\%
accuracy, ranking third in the list.

In a more controlled comparison---with shallow architectures and the basic interaction 
(linearly transformed and non-linearly squashed)---TBCNNs, of both variants, consistently outperform RNNs \cite{RAE}
to a large extent (50.4--51.4\% versus 43.2\%);
they also consistently outperform ``flat'' CNNs by more than 10\%.
Such results show that structures are important when modeling sentences;
tree-based convolution can capture these structural information more effectively than RNNs.

We also observe d-TBCNN achieves higher performance than c-TBCNN. This suggests that
compact tree expressiveness is more important than integrating global information
in this task.


\subsection{Question Classification}\label{ssQuestion}
We further evaluate TBCNN models on a question classification task.\footnote{
http://cogcomp.cs.illinois.edu/Data/QA/QC/} The dataset contains 5452 annotated sentences plus 500 test samples in TREC 10.
We also use the standard split, like \newcite{QC}.
Target labels contain 6 classes, namely
\verb"abbreviation", \verb"entity", \verb"description", \verb"human",
\verb"location", and \verb"numeric".
Some examples are also shown in Table \ref{tExample}.

\begin{table}
\centering
\small

\begin{tabular}{ccl}
\hline
\hline
\textbf{Method}      &\!\!\!\!\!\!\!\!\! \textbf{Acc. {\scriptsize(\%)}}\!\!\!\!\! & \textbf{Reported in}\\
\hline
\!\!\!SVM     &    \multirow{2}{*}{95.0}    &    \multirow{2}{*}{\newcite{QC}}  \\
  10k features + 60 rules & &\\
\hline
\!\!\!CNN-non-static          &  93.6       &  \newcite{cnn2}\\
\!\!\!CNN-mutlichannel        & 92.2      &   \newcite{cnn2}\\
\!\!\!RNN         & 90.2      &  \newcite{AdaCNN} \\
\!\!\!Deep-CNN    & 93.0      &  \newcite{CNNNLP}\!\!\! \\
\!\!\!Ada-CNN     & 92.4      &  \newcite{AdaCNN}  \\
\hline
\!\!\!c-TBCNN\!\!  & 94.8     & Our implementation \\
\!\!\!d-TBCNN\!\!  & \textbf{96.0}     & Our implementation          \\
\hline
\hline
\end{tabular}
\vspace{-.5cm}
\caption{\!Accuracy of 6-way question classification.\!\!}\label{tQC}

\vspace{-.5cm}
\end{table}

We chose this task to evaluate our models because the number of training samples is rather small,
so that we can know TBCNNs' performance when applied to datasets of different sizes.
To alleviate the problem of data sparseness,
we set the dimensions of convolutional layer and the last hidden layer to 30 and 25, respectively.
We do not back-propagate gradient to embeddings in this task.
Dropout rate for embeddings is 30\%; hidden layers are dropped out by 5\%.

Table \ref{tQC} compares our models to various other methods.
The first entry presents the previous state-of-the-art result,
achieved by traditional feature/rule engineering \cite{QC}. Their method
utilizes more than 10k features and 60 hand-coded rules.
On the contrary, our TBCNN models do not use a single human-engineered feature or rule.
Despite this, c-TBCNN achieves similar accuracy compared with feature engineering; d-TBCNN pushes the
state-of-the-art result to 96\%.
To the best of our knowledge, this is the first time that neural networks
beat dedicated human engineering in this question classification task.

The result also shows that both c-TBCNN and d-TBCNN reduce 
the error rate to a large extent, 
compared with other neural architectures in this task.

\subsection{Model Analysis}\label{ssModel}
In this part, we analyze our models quantitatively and qualitatively in several aspects,
shedding some light on the mechanism of TBCNNs.

\subsubsection{The Effect of Pooling}\label{sssPool}
The extracted features by tree-based convolution
have topologies varying in size and shape.
We propose in Section \ref{ssPool} several heuristics for pooling.
This subsection aims to provide a fair comparison among these pooling methods.

One reasonable protocol for comparison is to tune all hyperparameters for each setting and compare the
highest accuracy. This methodology, however, is too time-consuming, and
depends largely on the quality of hyperparameter tuning.
An alternative is to predefine a
set of sensible hyperparameters and report the accuracy under the same setting.
In this experiment, we chose the latter protocol,
where hidden layers are all 300-dimensional; no $\ell_2$ penalty is added.
Each configuration was run five times with different random initializations. We summarize 
the mean and
standard deviation in Table \ref{tPool}.

As the results imply, complicated pooling is better than global pooling to some degree
for both model variants.
But the effect is not strong; our models are not that sensitive to pooling methods, which mainly serve as a necessity for dealing with varying-structure data.
In our experiments, we apply 3-slot pooling for c-TBCNN and 2-slot pooling for d-TBCNN.

Comparing with other studies in the literature, 
we also notice that pooling is very effective and efficient 
in information gathering. 
\newcite{deepRNN} report 200 epochs for training a deep RNN, which achieves
49.8\% accuracy in the 5-class sentiment classification.
Our TBCNNs are typically trained within 25 epochs.
 
\subsubsection{The Effect of Sentence Lengths}\label{sssLength}
\begin{table}[!t]
\centering
\small
\begin{tabular}{ccc}
\hline
\hline
\textbf{Model} & \textbf{Pooling method} & \text{\textbf{5-class accuracy (\%)}}\\
\hline
\multirow{2}{*}{c-TBCNN} & Global  & 48.48 $\pm$ 0.54\\
        & 3-slot   & 48.69 $\pm$ 0.40\\
\hline
\multirow{2}{*}{d-TBCNN} & Global  & 49.39 $\pm$ 0.24\\
                         & 2-slot   & 49.94 $\pm$ 0.63\\
\hline
\hline
\end{tabular}

\vspace{-.2cm}
\caption{Accuracies of different pooling methods, averaged over
5 random initializations.
We chose sensible hyperparameters
manually in advance to make a fair comparison.
This leads to performance degradation (1--2\%) vis-a-vis Table \ref{tAcc}.}\label{tPool}
\end{table}

\begin{figure}[!t]
\centering
\vspace{-.3cm}
\includegraphics[width=.4\textwidth]{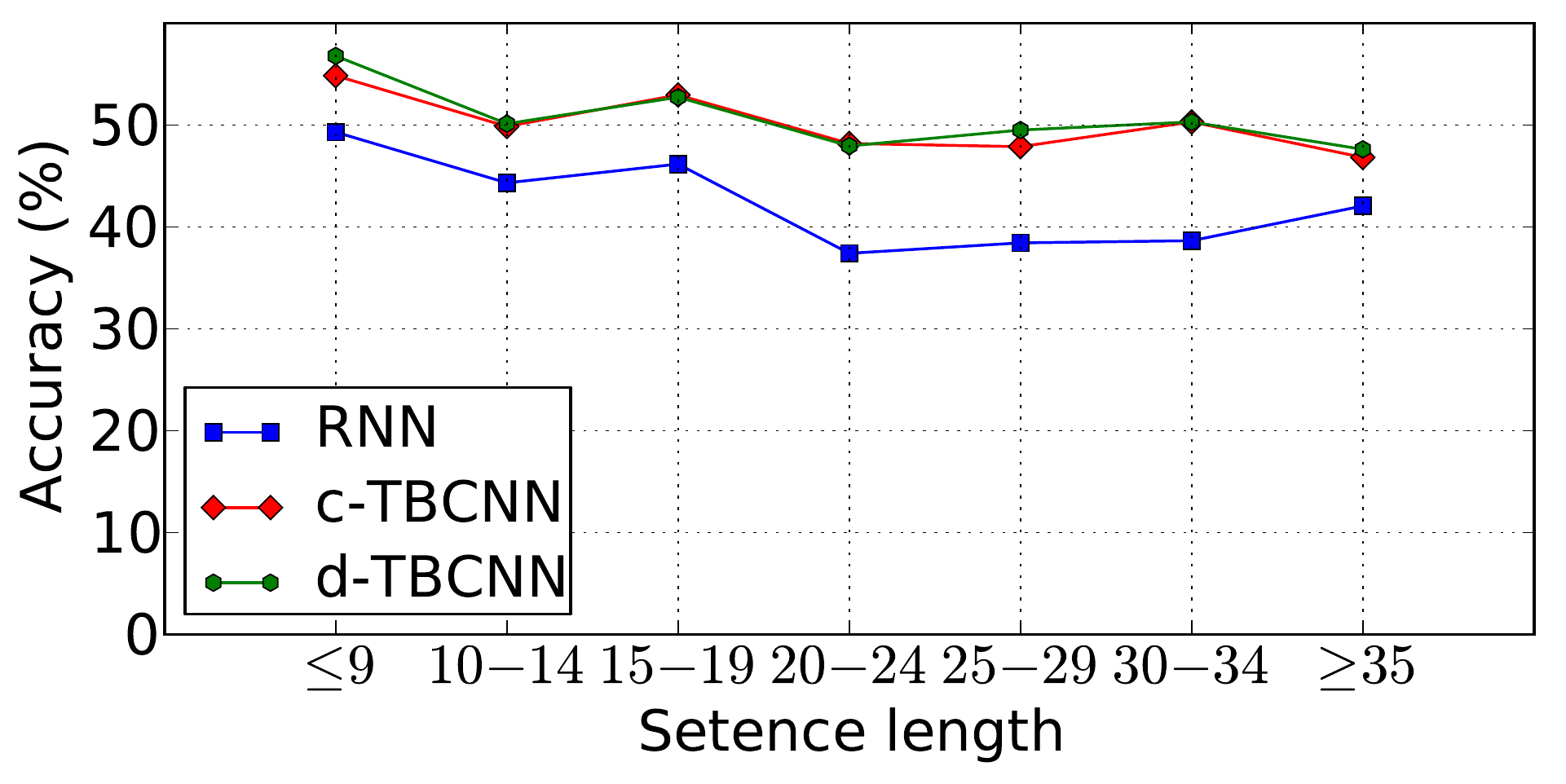}

\vspace{-.65cm}
\caption{Accuracies versus sentence lengths.}\label{fLength}
\vspace{-.5cm}
\end{figure}

\begin{figure*}[!t]
\centering

\vspace{-.5cm}

\includegraphics[width=.65\textwidth]{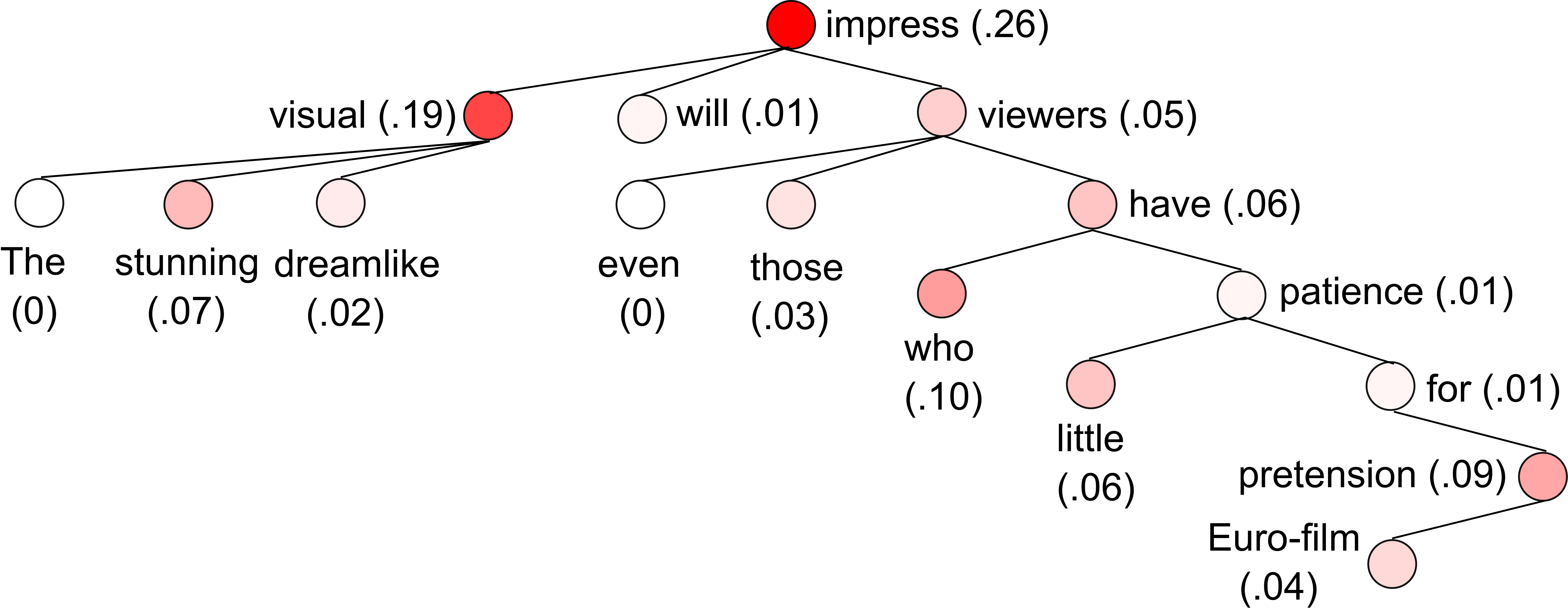}

\vspace{-.4cm}

\caption{Visualizing how features (after convolution) are related to the sentiment of a sentence. The sample corresponds a sentence in the dataset, ``The stunning dreamlike visual will impress even those viewers who have little patience for Euro-film pretension.''
The numbers in brackets denote the fraction
of a node's features  that are gathered by the max pooling layer
(also indicated by colors).}\label{fVisual}
\vspace{-.4cm}
\end{figure*}

We analyze how sentence lengths affect our models.
Sentences are split into 7 groups by length, with granularity 5.
A few too long or too short sentences are grouped together for smoothing;
the numbers of sentences in each group vary from 126 to 457.
Figure \ref{fLength} presents accuracies versus lengths
in TBCNNs.
For comparison, we also reimplemented RNN, achieving 42.7\% overall accuracy,
slightly worse than 43.2\% reported in \newcite{RAE}.
Thus, we think
our reimplementation is fair and that the comparison is sensible.

We observe that c-TBCNN and  d-TBCNN yield very similar behaviors.
They consistently outperform the RNN in all scenarios.
We also notice the gap, between TBCNNs and RNN, increases
when sentences contain more than 20 words.
This result confirms our theoretical analysis in Section \ref{sBackground}---for 
long sentences, the propagation paths in RNNs are deep, causing
RNNs' difficulty in information processing.
By contrast, our models explore structural information more effectively
with tree-based convolution. As information from any part of the tree
can propagate to the output layer with short paths,
TBCNNs are more capable for sentence modeling, especially for long sentences.

\subsubsection{Visualization}

Visualization is important to understanding
the mechanism of neural networks.
For TBCNNs, we would like to see how the extracted features (after convolution)
are further processed by the max pooling layer, and ultimately related to the supervised task.

To show this, we trace back where the max pooling layer's features come from.
For each dimension, the pooling layer chooses the maximum value from the nodes
that are pooled to it.
Thus, we can count the fraction in which a node's features
are gathered by pooling.
Intuitively, if a node's features are more related to the task,
the fraction tends to be larger, and vice versa.

Figure \ref{fVisual} illustrates an example processed by d-TBCNN in the task
of sentiment analysis.\footnote{
We only have space to present one example in the paper.
This example was not chosen deliberately. Similar traits can be found
through out the entire gallery, available on our website.
Also, we only present d-TBCNN, noticing that dependency trees are intrinsically more suitable for visualization
since we know the ``meaning'' of every node.}
Here, we applied global pooling
because information tracing is more sensible with one pooling slot.
As shown in the figure, tree-based convolution can effectively extract
information relevant to the task of interest.
The 2-layer windows corresponding to
``\textit{visual will impress viewers},'' ``\textit{the stunning dreamlike visual},'' say,
are discriminative to the sentence's sentiment. Hence, large
fractions (0.24 and 0.19) of their
features, after convolution, are gathered by pooling.
On the other hand,
words like \textit{the}, \textit{will}, \textit{even} are known as stop words \cite{stopwords}.
They are mostly noninformative for sentiment; hence, no (or minimal)
features are gathered. Such results are
consistent with human intuition.

We further observe that tree-based convolution does integrate information
of different words in the window.
For example, the word \textit{stunning} appears in two windows:
($a$) the window ``\textit{stunning}'' itself,
and ($b$) the window of ``\textit{the stunning dreamlike visual},'' with root node  \textit{visual},
\textit{stunning} acting as a child.
We see that Window $b$ is more relevant to the ultimate sentiment than Window $a$,
with fractions 0.19 versus 0.07,
even though the root \textit{visual} itself is neutral in sentiment.
In fact, Window $a$ has a larger fraction than the sum of its children's (the windows of ``\textit{the},'' ``\textit{stunning},'' and ``\textit{dreamlike}'').

\vspace{-.1cm}

\section{Conclusion}

\vspace{-.1cm}

In this paper, we proposed a novel neural
discriminative sentence model based on sentence parsing structures.
Our model can be built upon either constituency trees (denoted as c-TBCNN) or dependency trees (d-TBCNN).

Both variants have achieved high performance
in sentiment analysis and
question classification.
d-TBCNN is slightly better than c-TBCNN in our experiments, and 
has outperformed previous state-of-the-art results in both tasks. 
The results show that tree-based convolution
can capture sentences' structural information effectively, which
is useful for sentence modeling.

\bibliographystyle{acl}
\bibliography{dl}

\end{document}